\documentclass[sigconf,authorversion]{acmart}

\usepackage{amssymb}
\usepackage{amsmath}
\usepackage{float}
\usepackage{todonotes}
\usepackage{caption,subcaption}
\usepackage{makecell, diagbox}
\usepackage{multirow}
\usepackage{hhline}
\usepackage{color, colortbl}
\definecolor{Gray}{gray}{0.7}
\definecolor{sn1}{RGB}{184, 255, 198}
\definecolor{sn2}{RGB}{213, 204, 255}
\definecolor{sn3}{RGB}{255, 204, 209}
\usepackage{graphicx}
\graphicspath{{images/}}

\usepackage{soul}
\usepackage{ulem}
\usepackage{cancel}
\usepackage{comment}
\usepackage{xcolor}
\usepackage{amsmath}
\usepackage{adjustbox}
\usepackage{makecell}
\usepackage{rotating}
\usepackage{newtxmath}

\newcommand{\uls}{\begin{itemize}[leftmargin=*]}
\newcommand{\ule}{\end{itemize}}
\newcommand{\ols}{\begin{enumerate}[leftmargin=*]}
\newcommand{\ole}{\end{enumerate}}

\usepackage{enumitem}
\setlist[itemize]{leftmargin=*}
\setlist[enumerate]{leftmargin=*}
\setlist[description]{leftmargin=*}
\newcommand{\hlc}[2][yellow]{{\sethlcolor{#1} \hl{#2}}}

\newcommand{\para}[1]{\paragraph{\textnormal{\textbf{#1}:}}}

\usepackage{arydshln}
\usepackage{float}

\title{Unified Graph Networks (UGN): A Deep Neural Framework for Solving Graph Problems}
\author{Rudrajit Dawn}
\authornote{Work done during Master Thesis work at Jadavpur University.}
\email{rudrajit.dawn@gmail.com}
\affiliation{%
  \institution{TCS Research India}
  \city{Kolkata}
  \country{India}
}

\author{Madhusudan Ghosh}
\email{madhusuda.iacs@gmail.com}
\affiliation{%
  \institution{Indian Associantion for the Cultivation of Science}
  \city{Jadavpur, Kolkata}
  \country{India}
}

\author{Partha Basuchowdhuri}
\email{partha.basuchowdhuri@iacs.res.in}
\affiliation{%
  \institution{Indian Associantion for the Cultivation of Science}
  \city{Jadavpur, Kolkata}
  \country{India}
  }

\author{Sudip Kumar Naskar}
\email{sudip.naskar@gmail.com}
\affiliation{%
  \institution{Jadavpur University}
 \city{Jadavpur, Kolkata}
  \country{India}
}

\begin{document}
\setstcolor{red}




\begin{abstract}

Deep neural networks have enabled researchers to create powerful generalized frameworks, such as transformers, that can be used to solve well-studied problems in various application domains, such as text and image. However, such generalized frameworks are not available for solving graph problems. Graph structures are ubiquitous in many applications around us and many graph problems have been widely studied over years. In recent times, there has been a surge in deep neural network based approaches to solve graph problems, with growing availability of graph structured datasets across diverse domains. Nevertheless, existing methods are mostly tailored to solve a specific task and lack the capability to create a generalized model leading to solutions for different downstream tasks. In this work, we propose a novel, resource-efficient framework named \emph{U}nified \emph{G}raph \emph{N}etwork (UGN) by leveraging the feature extraction capability of graph convolutional neural networks (GCN) and 2-dimensional convolutional neural networks (Conv2D). UGN unifies various graph learning tasks, such as link prediction, node classification, community detection, graph-to-graph translation, knowledge graph completion, and more, within a cohesive framework, while exercising minimal task-specific extensions (e.g., formation of supernodes for coarsening massive networks to increase scalability, use of \textit{mean target connectivity matrix} (MTCM) representation for achieving scalability in graph translation task, etc.) to enhance the generalization capability of graph learning and analysis. We test the novel UGN framework for six uncorrelated graph problems, using twelve different datasets. Experimental results show that UGN outperforms the state-of-the-art baselines by a significant margin on ten datasets, while producing comparable results on the remaining dataset.

\end{abstract}

\maketitle

\section{Introduction}
\label{ss:intro}
Graphs are crucial for representing complex structures and relationships across diverse domains such as social network analysis (SNA), Internet of Things (IoT), biological networks, and transportation networks. The surge in graph data necessitates a generalized framework for graph analysis. This analysis is essential for tasks such as graph-to-graph translation, link prediction, and community detection. Graph-to-graph translation involves transforming an input graph into an output graph while preserving certain topological properties \cite{guo2019deep, lowe2014patent, van2013wu}. This process includes modifying the graph structure by adding, removing, or rearranging vertices and edges. Applications of graph-to-graph translation include graph transformation \cite{rong2020self}, graph synthesis \cite{nebli2020deep}, graph generation \cite{tang2020unbiased}, graph adaptation \cite{bai2020adaptive}, and drug discovery \cite{wu2022spatial}. Link prediction \cite{wang2020edge2vec, zitnikbiosnap} is also an important problem in the domain of graph analysis,  contributing to the understanding and utilization of graph data.

Some recent state-of-the-art (SOTA) approaches such as NEC-DGT~\cite{guo2019deep} and GT-GAN~\cite{9737289}, in graph-to-graph translation domain, have shown limitations in addressing the challenges of semi-supervised learning tasks, commonly employed in community detection task. These challenges include the resolution limit problem, where smaller communities within larger ones may remain undetected, and dynamic networks, where network structures evolve over time. 
Additionally, it does not perform well on the task of predicting brain functional connectivity (BFC)~\cite{van2013wu}. One prevalent explanation pertains to the intricate and dynamic nature of BFC~\cite{van2013wu} data samples, characterized by complex connectivity of brain structures that vary over time. The relationships between nodes and edges are shaped by neural activity, the functions of brain regions, and external stimuli. This intricate nature poses a challenge for simplistic NEC-DGT~\cite{guo2019deep} frameworks in capturing the complex topological structure.

Another extensively researched area in the field of graph analysis is the study of link prediction, focusing on tasks like anticipating trust or friendship connections within social networks \cite{wang2020edge2vec, Tang-etal12c, leskovec2009community}. Various established approaches in this domain are edge2vec~\cite{wang2020edge2vec}, MEIM~\cite{DBLP:conf/ijcai/TranT22}, HOGCN~\cite{kishan2021predicting}, GHRS~\cite{darban2022ghrs}. However, these frameworks are not tailored for graph-to-graph translation problems, where specific nodes or edges from the source graph might not be present in the target graph. While these techniques ~\cite{wang2020edge2vec,kishan2021predicting} demonstrate proficiency in link prediction by effectively capturing intricate relationships and structures within a single graph, leveraging both local and global features to predict missing connections, they ~\cite{wang2020edge2vec,darban2022ghrs} fall short in the context of graph-to-graph translation, which necessitates more intricate modifications, demanding a generalized model to comprehend and adjust the graph's overall structure.

To alleviate these challenges, where most of the existing frameworks do not perform well for different tasks on various domains, we propose a generalized framework namely UGN~\footnote{Our source code is available here: https://anonymous.4open.science/r/UGN} which performs significantly well on different downstream tasks present in the existing literature of graph learning and analysis.

We make the following contributions in the paper.
\begin{enumerate}
\item We propose a novel encoder-decoder based framework to combine the strengths of neural message passing and convolutional neural networks for various graph analysis tasks. Empirical findings demonstrate that our proposed framework, with minimal modifications, consistently achieves excellent performance, yielding  state-of-the-art (SOTA) or comparable results across various downstream tasks in both supervised and semi-supervised settings.

\item Additionally, we propose a novel strategy to initialize node feature representation for processing large graphs in low resource environment.
\end{enumerate}

\section{Related Works}

Since our work mainly investigates deep metric learning based graph analysis tasks, we first discuss about the SOTA techniques available in graph learning literature. Later, we differentiate our proposed framework from the existing ones.

\citet{guo2019deep} proposed a generic, end-to-end framework for joint node and edge attributes prediction, which is a need for real world applications such as malware confinement in IoT networks and structural-to-functional network translation. \citet{bruna2014spectral} first introduced the spectral graph convolutional neural networks, and then it was extended by \citet{defferrard2016convolutional} using fast localized convolutions, which is further approximated by~\citet{DBLP:conf/iclr/KipfW17} for an efficient architecture in a semi-supervised setting. \citet{li2018diffusion} proposed a Diffusion Convolution Recurrent Neural Network (DCRNN) for traffic forecasting which incorporates both spatial and temporal dependency in the traffic flow. \citet{Yu2018SpatioTemporalGC} formulated the node attributes prediction problem of graphs based on the complete convolution structures. \citet{9737289} addressed graph topology translation using a generative model with graph convolution, deconvolution layers, and a novel conditional graph discriminator. \citet{wang2020edge2vec} proposed edge-based graph embedding (edge2vec) to map the edges in social networks directly to low-dimensional vectors by preserving the necessary topological properties of networks. Additionally, \citet{perozzi2014deepwalk} proposed DeepWalk by utilizing the truncated random walks to obtain the node level information. Later on, Node2vec algorithm was proposed by \citet{grover2016node2vec} to improve the DeepWalk by replacing the truncated random walks with a combination of Breadth First Search (BFS) and Depth First Search (DFS) algorithms. Another direction of this work is that \citet{tang2015line} proposed LINE to represent the contextual representation of the nodes present in a Graph, while \citet{cao2015grarep} extended it with GraRep to include indirect neighbors, facing issues with embedding vector length. Discriminative Deep Random Walk (DDRW)~\cite{li2016discriminative} and max-margin DeepWalk (MMDW)~\cite{tu2016max}, respectively, optimize classifier and embedding jointly. Additionally, RSDNE~\cite{wang2018rsdne} produced a comparable performance in both general and zero-shot scenario where graph embedding by ensures the intra-class similarity and inter-class dissimilarity. The works by \citet{newman2006finding} and \citet{newman2013spectral} serve as examples of spectral methods. Statistical inference based approaches typically employed conventional methods to fit data to a generative network model. A commonly used generative model for  network communities is the Stochastic Block Model (SBM). \citet{clauset2005finding} and~\citet{lancichinetti2011limits} utilizes modularity based optimization technique for community detection task whereas \citet{reichardt2006statistical}, and \citet{rosvall2008maps} utilizes random walks, diffusion based algorithm to preserve the community structure.

Additionally, knowledge graph embedding is an active research area with many different methodologies. This task can be categorized based on the interaction mechanisms and the computational matching score. Tensor-decomposition-based models adapt tensor representations in CP format, Tucker format, and block term format to represent the knowledge graph data~\citep{kolda2009tensor} by proposing SOTA models such as ComplEx~\citep{trouillon2016complex}, SimplE~\citep{kazemi2018simple}, TuckER (\citep{balavzevic2019tucker}), and MEIM (\citet{DBLP:conf/ijcai/TranT22}). They often achieve good trade-off between efficiency and expressiveness. Recently, ConvE~\citep{dettmers2018convolutional}, CompGCN~\citep{vashishthcomposition} have been investigated by utilizing the neural network to embed the knowledge graph data. Translation-based models use geometrical distance to compute the matching score, with relation embeddings such as TransE (\citet{bordes2013translating}).  There are several ways to use orthogonality in knowledge graph embedding, such as RotatE (\citet{sunrotate}) using complex product, Quaternion (\citet{tran2019analyzing}) and QuatE (\citet{zhang2019quaternion}) using quaternion product, GCOTE (\citet{tang2020orthogonal}) using the Gram Schmidt process, and RotH (\citet{chami2020low}) using the Givens matrix.


Our model is unique in terms of scalability and genericness:
(i) We experimented with smaller graphs containing 20-60 nodes as well as large graphs containing 20K-80K nodes using the same model architecture and obtained SOTA or near-SOTA results in both cases.
(ii) Our model can smoothly transit from supervised learning domain to semi-supervised learning domain only by adding an extra component in the loss function, keeping the model architecture same. We applied our model for supervised tasks where number of known labels of nodes or edges in the training set is much larger or at least similar to the test set, and semi-supervised tasks where number of known labels in the training set is much less than the test set, producing SOTA or near-SOTA results for both the tasks.

\section{Model Architecture}


\subsection{Encoder-Decoder Model}
\label{EandD}
\begin{figure}[t]
\centering
\vspace{-.2cm}
\includegraphics[width=0.5\textwidth]{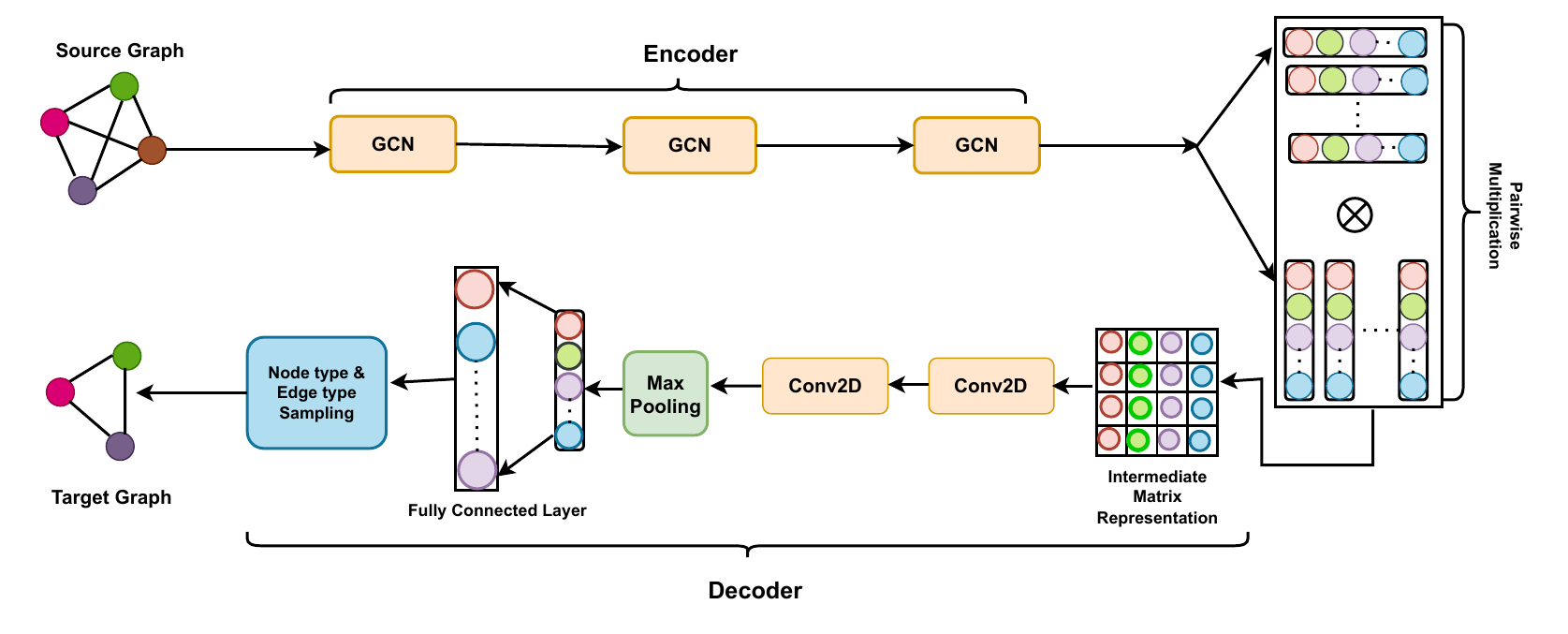}
\caption{
\small Pictorial overview of our UGN framework.
}
\label{fig:arch}
\end{figure}

The proposed model uses an encoder-decoder architecture. The encoder transforms the input into a dense vector that captures key features, typically with lower dimensions to reduce storage and computational costs. The decoder then uses this dense representation to generate the desired output in the appropriate format.The overview of our proposed framework namely UGN has been depicted in Figure~\ref{fig:arch}.

\para{Encoder}
\label{para:enc}
The encoder module uses a sequence of Graph Convolutional Network (GCN) \cite{4700287} layers to encode meaningful representations of nodes. It takes the input feature matrix and the adjacency matrix as input and generates the latent space representation of the nodes as output. The input graph $G$ has $N$ nodes and $C$ number of features for every node. Therefore, the input feature matrix $F$, which stores feature vectors of every node as its row, is of size $N \times C$.
Adjacency matrix of $G$ is $A$, with dimensions $N \times N$. Let, $I$ be an identity matrix of size $N \times N$. 

 New adjacency matrix $\widetilde{A}$ introduces self-loop to all the nodes in the graph, i.e., $\widetilde{A} = A + I$, to avoid self-information loss in the message passing operation.
Suppose, $D$ is the degree matrix with all non-diagonal elements being zero and diagonal elements representing degree of the nodes i.e., $D_{i,i}$ measures the degree of $i^{th}$ node including self-connections.

\vspace*{-\baselineskip}
\begin{equation}
\widehat{A} = D^{-\frac{1}{2}}\times \widetilde{A} \times D^{-\frac{1}{2}}
\end{equation}

If the output feature matrix $F_{h_1}$ is of size $N \times H_1$ and the trainable weight matrix of the GCN layer, $W^{(0)}$, is of size $C \times H_1$, then $F_{h_1} = ReLU(\widehat{A} \times F \times W^{(0)})$.
Similarly, for a subsequent GCN layer, which takes the the edge index and the output feature matrix $F_{h_1}$ from the previous layer as input and generates output feature matrix $F_{h_2}$ of size $N \times H_2$ using trainable weight matrix $W^{(1)}$ of size $H_1 \times H_2$, we can state,
\begin{equation}
\begin{split}
F_{h_2} & = ReLU(\widehat{A} \times F_{h_1} \times W^{(1)})\\
& = ReLU(\widehat{A} \times ReLU(\widehat{A} \times F \times W^{(0)}) \times W^{(1)})
\end{split}
\end{equation}
There are $k$ GCN layers in the encoder. Consequently, the final latent space representation of nodes is given by $F_{h_k}$ of size $N \times H_k$, where $H_k$, represented as $L$ in the rest of the paper, is the latent dimension. These latent representations are passed to the decoder block for solving node and edge classification tasks.

\para{Decoder}
\label{par:decoder}
The decoder consists of two dimensional convolutional layers\cite{fukushima1980neocognitron}, max-pooling layers and linear layers. Graph problems may require the decoder to predict whether an edge $e$ incident on nodes $u$ and $v$ exists (link prediction), or predicting the class label attached to it (edge classification). For solving such problems, let the latent vectors for the nodes $u$ and $v$ be $l_u$ and $l_v$, respectively.
\begin{equation}
\begin{split}
l_u & = \left[ F_{h_k(u,0)},  F_{h_k(u,1)}, \dots ,  F_{h_k(u,L-1)}\right]
= \left[ u_0, u_1, \dots , u_{L-1}\right]\\
l_v & = \left[ F_{h_k(v,0)},  F_{h_k(v,1)}, \dots ,  F_{h_k(v,L-1)}\right]
= \left[ v_0, v_1, \dots , v_{L-1}\right]
\end{split}
\end{equation}
An intermediate matrix $M$ of size $L \times L$ is formed 
following equation (\ref{eqn:intm}) 
using the products of all pairs of elements, where $M_{ij}$ = $v_j \times u_i$, $i$, $j$ are integers in [0, $L$-1]. The pairwise multiplication allows all possible latent feature pairs from the source node and target node to be considered as input, while concatenation only allows to consider them separately. In terms of our model, pairwise multiplication helps revealing inter-feature dependency or relationship between the source nodes and target nodes, whereas concatenation reveals inter-class dependency. 
Let the model parameters be $\Theta$ and output be $u$ and $v$. While training the model, if $u$ and $v$ are computed separately, then their individual losses can be back-propagated as $\frac{\partial \left(u+v\right)}{\partial \Theta}=\frac{\partial \left(u\right)}{\partial \Theta} + \frac{\partial \left(v\right)}{\partial \Theta}$. Instead, if $u\cdot v$ is computed, then each loss component is supported by the counterpart while back-propagating as $\frac{\partial \left(u\cdot v\right)}{\partial \Theta}=u\cdot\frac{\partial \left(v\right)}{\partial \Theta} + v\cdot\frac{\partial \left(u\right)}{\partial \Theta}$ .

\vspace*{-\baselineskip}
\begin{equation}
M=l_v^T \times l_u
\label{eqn:intm}
\end{equation}

The internal feature representation, denoted by $M\in \mathbb{R}^{L \times L}$, is considered as a one-channel $L \times L$ image to enhance the performance of the downstream classification tasks. The image-like representation of $M$ allows for the utilization of convolution and max-pooling operations to generate significant abstract representations. To achieve this, the internal input representation ($M$) is passed into the required convolution layer, followed by the max-pooling operation, as we treat $M$ as an image. To obtain the most crucial abstract feature representation, we repeat the convolutional and max-pooling operations for $s$ times. The abstract output representation of the max-pooling layer is subsequently passed into the $t$ linear layers followed by the necessary activation function to predict the required edges and their corresponding types.
The values of $k$, $s$ and $t$ range from three to five depending on the size of the graph and input feature dimension of the nodes. While working with node classification problem, the intermediate matrix $M$ is formulated by replacing $l_v$ with $l_u$ in the equation (\ref{eqn:intm}) for classifying node $u$. Finally, we train our model in an end-to-end fashion, using categorical cross-entropy loss function as shown in equation (\ref{eqn:sup_loss}), where $\mathcal{L}_s$ is the supervised loss, $V$ is the node set, $c$ is the number of classes, $V=V_k \cup V_u$, where $V_k$ is the set of nodes with known node-labels and $V_u$ is the set of nodes with unknown node-labels (generally, $V_u=\emptyset$), $N$, $N_k$ and $N_u$ are the number of nodes in the node sets $V$, $V_k$ and $V_u$, respectively. Furthermore, $T_{N\times c}$ is the one-hot encoded ground-truth labels of all the nodes, $S_{N\times c}$ is the logits of all the nodes, and $O_{N\times c}$ is the transformer output for all the nodes. Similarly, for edge classification, $V$ is replaced by $E$ in the equation (\ref{eqn:sup_loss}), where $E$ is the edge set.
\vspace{-.2cm}
\begin{equation}
\begin{split}
\mathcal{L}_s &= -\sum_{v\in V_k}\sum_{i=0}^{c-1}T[v][i]\times\log (S[v][i])\\
&= -\sum_{v\in V_k}\sum_{i=0}^{c-1}T[v][i]\times\log\left(\frac{e^{O[v][i]}}{\sum_{j=0}^{c-1}e^{O[v][j]}}\right)
\end{split}
\label{eqn:sup_loss}
\end{equation}


\subsection{Special Cases of our Proposed Approach}
\label{sec:SpecialCases}
In this section, we investigate whether integration of diverse components into our proposed model can facilitate its applicability to a wide range of graph related problems.
\subsubsection{Supernode Feature}
\label{super}

Many existing datasets (e.g., Slashdot, BioSNAP-DDI, BioSNAP-DTI, etc.) used for solving graph problems generally suffer from the absence of node features. The prevalent approaches \cite{wang2020edge2vec,kishan2021predicting}, initialize such features with one-hot representation of the nodes. However, when the size of the graph is substantially large, using one-hot encoding as initial node feature becomes unsuitable due to high memory requirement. To mitigate this problem, we introduce a novel component, namely \textit{supernode} feature, to accommodate large graphs into neural network frameworks in low-resource environment. Supernodes are mutually exclusive and exhaustive subsets of the node set $V$. Suppose, a graph $G$ contains $N$ number of nodes and $s$ number of supernodes.
If $N$ is divisible by $s$, then each supernode consists of exactly $k=\frac{N}{s}$ number of nodes. Otherwise, some supernodes (say, $i$ many) have $k$ nodes while the rest have $m$ nodes, such that $N=k\times i+m\times (s-i)$. Let, $S_j$ be a supernode set, where $j$ is an integer in [0, s-1]. For each node $v \in V$, we compute a connection vector $d^i=[d^v_0,d^v_1,…,d^v_{s-1}]$, where $d^v_j$ is the normalized number of connections between $v$ and $S_j$, i.e.,\\ $d^v_j$ = $\frac{\text{\# of connections between $v$ and $S_j$}}{ \text{\# of nodes in $S_j$}}$. Given the graph is undirected, this vector captures the connection strength of a node to each supernode. 

Although many graph coarsening methods are known in classical graph problems, such techniques have not been popularly used in neural network frameworks. In our model, we apply coarsening of large graphs through the supernode feature to alleviate memory requirement and empirically show that it contributes to additional performance gain.



\subsubsection{Unsupervised Loss for Semi-supervised Task}

\label{unsupervised}
A popular challenge in neural graph analysis is posed by the semi-supervised task of community detection, where only a few nodes are labeled with community labels \cite{zachary1977information}. In this scenario, a supervised classification loss, when employed in conjunction with encoder-decoder framework, fails to capture adequate structural information of the entire graph. To leverage the efficacy of the unlabeled nodes, we introduce an unsupervised loss function ($\mathcal{L}_s$) that supplements the supervised loss function ($\mathcal{L}_s$, as in equation (\ref{eqn:sup_loss})).
The unsupervised component of the loss function, $\mathcal{L}_u$, is defined as in equation (\ref{eqn:unsup_loss}), where $c$ is the number of classes, $O_{N\times c}$ is the model output of all the nodes, and $E$ is the edge set of the graph. 

\vspace*{-\baselineskip}
\begin{equation}
\mathcal{L}_u = \sum_{(u,v) \in E}\sum_{i=0}^{c-1}\frac{1}{c}\times(O[u][i]-O[v][i])^2
\label{eqn:unsup_loss}
\end{equation} 
\vspace{-.3cm}

The loss function has been designed based on the assumption that densely connected nodes tend to belong to the same community. 
This assumption leads to the expectation that MSE-loss between the transformer-outputs for two connected nodes should converge to zero. The final loss function $\mathcal{L}$ is defined as $\mathcal{L} = \mathcal{L}_s + \mathcal{L}_u$. 

The problem of community detection can be compared to an environment where nodes are balls in a multidimensional space, connected via springs in place of edges. In this setup, the nodes with community labels have fixed positions. Here, the unsupervised loss $\mathcal{L}_u$ is the potential energy stored in the springs. The model tries to minimize this energy to reach equilibrium.


\subsubsection{Mean Target Connectivity Matrix (MTCM) Rrepresentation for Complete Graphs}
\label{mtcm}
In the graph translation task, it is generally observed that when both the source and target graphs are complete, the corresponding adjacency matrices are populated by a correlation-based connectedness score. This score ranges from -1 to 1 and is assigned to all possible pairs of nodes in the graph. The adjacency matrix for the source graph is referred to as the source connectivity matrix ($S$) representation while that of the target graph is known as the target connectivity matrix ($T$). Being complete graphs, the number of edges escalates significantly to the maximum possible connections between the nodes, which imparts a direct effect on the training duration of the model. To address this issue and leverage the similarity present in the target connectivity matrices, we propose a mean target connectivity matrix (MTCM) representation component to reduce the computational resources. The MTCM is constructed using equation (\ref{eqn:mtm}) where $\overline{T}$ represents the required MTCM, $N$ denotes the order of the graphs in the dataset, and $|D|$ corresponds to the total number of source-target connectivity pairs within the dataset.


\vspace*{-\baselineskip}
\begin{equation}
\overline{T}[i][j]=\frac{1}{\mid D\mid}\sum_{T\in Dataset}T[i][j],\forall i,j\in[0,N-1]
\label{eqn:mtm}
\end{equation}

Subsequently, for each target connectivity matrix, a difference matrix $Dif$ is created by subtracting MTCM from the corresponding target connectivity matrix as in equation (\ref{eqn:diff_matrix}), where $Dif_n$ denotes the difference matrix for the target connectivity matrix $T_n$. The model is trained in an end-to-end fashion to predict $Dif_n$ by taking source connectivity matrix (i.e., $S_n$) as the required input representation.

\vspace*{-\baselineskip}
\begin{equation}
Dif_n[i][j]=T_n[i][j]-\overline{T}[i][j], \forall i,j\in[0,N-1], \forall n\in[0,\mid D\mid -1]
\label{eqn:diff_matrix}
\end{equation}

The predicted connectivity matrix is obtained by adding the MTCM to the predicted difference matrix from a source connectivity matrix. Using this method, the model can exploit the similarity among the target connectivity matrices in its prediction and helps improve the evaluation scores.

\section{Datasets}
\label{s:dataset}
\label{dataset} We conduct our experiments on various datasets related to different graph learning tasks, as described below.

\begin{enumerate}
\item \label{data:iot}\textbf{IoT} \cite{guo2019deep}: This dataset is used for malware confinement tasks. It includes three subsets with graph orders of 20, 40, and 60, and average node degrees of 7.42, 11.22, and 9.51, respectively. Each subset contains 343 graph pairs with contextual parameters: infection rate, recovery rate, and decay rate. Node attributes indicate infection status, and edge attributes represent the distance between nodes.

\item \label{data:chem}\textbf{Chemistry Reaction} \cite{guo2019deep, lowe2014patent}: This dataset is used to forecast the outcome of a chemical reaction based on the reactants. Molecular graphs have atoms as nodes and bonds as edges. There are 7,180 atom-mapped reactant-product pairs in the dataset with an average node degree of 1.75. Atom characteristics consist of elemental identity, connectivity, hydrogen count, valence, and aromaticity. Moreover, bond characteristics encompass bond type and connectivity.



\item \label{data:socailnet}\textbf{Social Networks}: We have selected two datasets used for directed link prediction task.
\begin{enumerate}
\item \label{data:epi}\textbf{Epinions}
\cite{wang2020edge2vec}\cite{Tang-etal12c}: It is a large social network with 22,166 nodes and 355,217 directed edges. Here, nodes are users and edges represent trust between users. Users provided ratings (1-5) of different products under 27 categories and each rating has a helpfulness score (1-5). 
\item \label{data:slash}\textbf{Slashdot}
\cite{wang2020edge2vec}\cite{leskovec2009community}: It is another large social network graph with 82,168 nodes and 870,161 directed edges. Here, nodes are users and edges represent friendship between users.
\end{enumerate}

\item \label{data:community}\textbf{Community Detection}: In this task, label of only one node from each community is known during training, making it a semi-supervised task. Here, edges represent interactions between nodes.
\begin{enumerate}
\item \label{data:zach}\textbf{Zachary’s karate club} \cite{zachary1977information}\cite{lu2018community}\cite{DBLP:journals/debu/HamiltonYL17}\cite{cai2019community}: It is a small bidirectional graph with 34 nodes and 156 edges. Some researchers (on the basis of their notion of community) have reported the optimal number of communities to be two, whereas others have reported it to be four.
\item \label{data:acf}\textbf{American College football} \cite{girvan2002community}\cite{lu2018community}\cite{cai2019community}: It is another small bidirectional graph with 115 nodes, 1,226 edges and 12 communities. The community sizes are non-uniform.
\item \label{data:books}\textbf{Books about US politics} \cite{krebs2004books}\cite{cai2019community}: It is also a small bidirectional graph with 105 nodes, 882 edges and 3 communities (i.e., liberal, conservative and neutral). Collection size under neutral category is much less than the other two categories.
\end{enumerate}

\item \label{data:biomed}\textbf{Biomedical}:
\begin{enumerate}
\item \label{data:hcp}\textbf{Human Connectome Project (HCP\footnote{https://compneuro.neuromatch.io/projects/fMRI/README.html})} \cite{van2013wu}: The dataset comprises 339 samples with structural (SC) and functional (FC) brain connectivity data across five cognitive tasks (rest, gambling, language, motor, emotion). Each network consists of 360 nodes that correspond to distinct brain regions, featuring adjacency matrices based on correlation. The objective is to forecast functional connectivity (FC) utilizing structural connectivity (SC).

\item \label{data:ddi}\textbf{Drug Drug Interaction (BioSNAP-DDI\footnote{http://snap.stanford.edu/biodata/datasets/10001/10001-ChCh-Miner.html})} \cite{zitnikbiosnap}: This dataset contains interactions between 1,514 FDA-approved drugs, with 48,514 bidirectional edges. The task is to predict drug interactions (edges).

\item \label{data:dti}\textbf{Drug Target Interaction (BioSNAP-DTI\footnote{https://snap.stanford.edu/biodata/datasets/10002/10002-ChG-Miner.html})} \cite{zitnikbiosnap}: This dataset includes interactions between 5,017 drugs and 2,324 target proteins, with 15,138 bidirectional edges. The task is to predict drug-target interactions.

\item \label{data:ppi}\textbf{Protein Protein Interaction (HuRI-PPI\footnote{http://www.interactome-atlas.org/data/HuRI.tsv})} \cite{luck2020reference}: This dataset contains interactions between 8,245 proteins with 52,068 bidirectional edges. The task is to predict protein-protein interactions.
\end{enumerate}

\item \label{data:yago}\textbf{YAGO} \cite{suchanek2007yago}: YAGO\footnote{https://yago-knowledge.org/} is a Knowledge Graph that adds common knowledge facts from Wikipedia to WordNet, transforming it into a common knowledge base. It is a large directed knowledge graph where nodes represent entities and directed edges describe facts between them. YAGO includes 37 relation types and each entity has at least 10 relations. The knowledge graph contains 123,131 entities and 1,089,040 relations between them.

\end{enumerate}



\section{Experiments and Results}
\label{ss:res_analysis}


In this section, we present the experiments conducted on various datasets, detailing task-specific and dataset-specific experimental setups, the corresponding results, and a comparison of performance between our model and the state-of-the-art (SOTA) models.


\subsection{IoT}
The malware confinement task consists of two subtasks: predicting the infection state of the nodes (interconnected devices) and restructuring of the edges i.e., connection between devices. To conduct the experiment, we initialize the node features with either 1 or 0 values, as $F_{malicious} = [1, 1, \dots , 1]_{1\times N}$ and $F_{benevolent} = [0, 0, \dots , 0]_{1\times N}$, 1 and 0 denoting the malicious and benevolent nodes respectively, where $N$ is the order of the graph. 

Let the latent representation of a particular node $v_i$ from the produced output representation of the encoder block be represented as $l_i=[l_{i_0}, l_{i_1}, \dots , l_{i_{L-1}}]$ where $L$ is the latent dimension. Before passing this vector to the decoder as input, its inverse distance vector $d_{i}^{-1}$ is concatenated with it, where $d_i=[d_{i0}, d_{i1}, \dots , d_{i(N-1)}]$ and $d_{ij}$ is the distance between $v_i$ and $v_j$ . Let the new latent feature for $v_i$ be $l_i^{new}=\left[ l_i,d_i^{-1} \right]$. The inverse distance vector gives the local structural information of a node. Furthermore, $F_{malicious}$ and $F_{benevolent}$ are available as node features. It is given to the decoder since distance between nodes does not change throughout the process; otherwise it could be given to the encoder. Additionally, to incorporate self-information in the feature representation of that node, we introduce self-loops with distance 1 and the required resultant input matrix ($M$)  of size $(L+N)\times (L+N)$ is computed using equation \ref{eqn:iotmi}, and subsequently passed into the decoder for classifying $v_i$ as well as to classify the edges. It is to be noted that edge classification does not always need separate intermediate representation. For the IoT dataset, the node representations serve both the tasks of node classification and edge classification which reduces computational costs, unlike the Chemistry Reaction dataset where separate node and edge representations are needed. The inverse distance vector is used as connectedness vector where 0 signifies no connection and 1 signifies the node itself, rest of the values lie between 0 and 1. 
$F_{malicious}$ and $F_{benevolent}$ are taken as $N$ dimensional to give same importance to the features as the structural information.

\begin{equation}
\begin{split}
M_i &= \left(l_i^{new}\right)^T \times l_i^{new}
=\left[ l_i,d_i^{-1} \right]^T \times \left[ l_i,d_i^{-1} \right]\\
&=\begin{bmatrix}
l_i^T \times l_i & l_i^T \times d_i^{-1}\\
\left(d_i^{-1}\right)^T \times l_i & \left(d_i^{-1}\right)^T \times d_i^{-1}
\end{bmatrix}
\end{split}
\label{eqn:iotmi}
\end{equation}



For each subset (with graph order of 20, 40 and 60) of the IOT dataset, 
50 data points (i.e., graph-pairs) are used as training set with batch size 1 and learning rate 0.005, and the rest (293 graph-pairs) is used as the test set. 

\begin{table*}[htbp]
    \centering
    \begin{subtable}{.5\linewidth}
      \centering
      \scalebox{0.8}{
        \begin{tabular}{|c|c|c|c|c|c|c|}
        \hline
        & \multicolumn{2}{c|}{IoT--20} & \multicolumn{2}{c|}{IoT--40} & \multicolumn{2}{c|}{IoT--60} \\ 
        \hline
        Method & \makecell{Edge\\Acc.} & \makecell{Node\\Acc.} & \makecell{Edge\\Acc.} & \makecell{Node\\Acc.} & \makecell{Edge\\Acc.} & \makecell{Node\\Acc.}\\ [0.5ex]
        \hline
        GraphRNN \cite{you2018graphrnn} & 0.61 & -- & 0.71 & -- & 0.84 & -- \\
        IN \cite{battaglia2016interaction} & -- & 0.88 & -- &  0.88 & -- & 0.87 \\
        GraphVAE \cite{simonovsky2018graphvae} & 0.51 & -- & 0.61 & -- & 0.81 & -- \\
        DCRNN \cite{li2018diffusion} & -- & 0.88 & -- &  0.88 & -- & 0.87 \\
        GT-GAN \cite{9737289} & 0.63 & -- & 0.90 & -- & 0.95 & -- \\
        STGCN \cite{Yu2018SpatioTemporalGC} & -- & 0.92 & -- & 0.93 & -- & \textbf{0.94} \\
        NR-DGT \cite{guo2019deep} & 0.91 & 0.91 & 0.91 & 0.89 & 0.95 & 0.88 \\
        NEC-DGT \cite{guo2019deep} & 0.92 & \textbf{0.93} & 0.94 & 0.93 & 0.96 & 0.90 \\
        UGN & \textbf{0.93} & 0.92 & \textbf{0.96} & \textbf{0.94} & \textbf{0.98} & \textbf{0.94} \\ [0.5ex]
        \hline
        \end{tabular}}
        \caption{IoT}
        \label{tab:IoT_compare}
    \end{subtable}%
    \begin{subtable}{.2\linewidth}
      \centering
      \scalebox{0.85}{
        \begin{tabular}{|c|c|}
        \hline
        Method & \makecell{Edge\\Acc.} \\ [0.5ex]
        \hline\hline
        GT-GAN \cite{9737289} & 0.87 \\ [0.5ex]
        WLDN \cite{jin2017predicting} & 0.97 \\ [0.5ex]
        NR-DGT \cite{guo2019deep} & \textbf{0.99} \\ [0.5ex]
        NEC-DGT \cite{guo2019deep} & \textbf{0.99} \\ [0.5ex]
        UGN & 0.98 \\ [0.5ex]
        \hline
        \end{tabular}}
        \caption{Chemistry}
        \label{tab:Chem_compare}
    \end{subtable}%
    \begin{subtable}{.3\linewidth}
      \centering
      \scalebox{0.85}{
        \begin{tabular}{c|c|c|c|}
        \cline{2-4}
        \multicolumn{1}{l|}{}        & \multicolumn{3}{c|}{Node Acc.}\\ \cline{1-4}        
        \multicolumn{1}{|c|}{Method} & \rotatebox{90}{\makecell{Karate\\Club}} & \rotatebox{90}{Football} & \rotatebox{90}{\makecell{US Politics\\Books}}\\
        \hline
        \multicolumn{1}{|c|}{GN \cite{girvan2002community}}  & 0.97 & 0.80 & 0.81\\
        \multicolumn{1}{|c|}{FG \cite{clauset2004finding}} & 0.97 & 0.53  & 0.73\\
        \multicolumn{1}{|c|}{MIGA \cite{shang2013community}} & \textbf{1.00} & 0.86 & 0.80\\
        \multicolumn{1}{|c|}{SLC \cite{mahmood2015subspace}} & 0.97 & 0.85 & 0.80\\ 
        \multicolumn{1}{|c|}{Eqn (20) \cite{tian2018community}}  & \textbf{1.00} & 0.86 & 0.83\\
        \multicolumn{1}{|c|}{$k$-means \cite{cai2019community}} & 0.88 & 0.73 & 0.66\\
        \multicolumn{1}{|c|}{DDJKM \cite{cai2019community}} & \textbf{1.00} & 0.89  & 0.73\\
        \multicolumn{1}{|c|}{UGN} & \textbf{1.00} &  \textbf{0.92}  & \textbf{0.88}\\
        \hline
        \end{tabular}}
        \caption{Community Detection}
        \label{tab:ComDec_compare}
    \end{subtable}%
    
    \begin{subtable}{.3\linewidth}
      \centering
      \scalebox{0.85}{
        \begin{tabular}{c|c|c|}
        \cline{2-3}
        \multicolumn{1}{l|}{}        & \multicolumn{2}{c|}{Edge Acc.}\\ \cline{1-3}
        
        \multicolumn{1}{|c|}{Method} & Epinions & Slashdot \\
        \hline
        \multicolumn{1}{|c|}{LINE \cite{tang2015line}}  & 0.65 & 0.70 \\
        \multicolumn{1}{|c|}{SDNE \cite{wang2016structural}} & 0.71 & 0.69 \\
        \multicolumn{1}{|c|}{DeepWalk \cite{perozzi2014deepwalk}} & 0.72 & 0.70\\
        \multicolumn{1}{|c|}{DeepEdge \cite{abu2017learning}} & 0.81 & 0.77\\ 
        \multicolumn{1}{|c|}{node2vec \cite{grover2016node2vec}}  & 0.80 & 0.75\\
        \multicolumn{1}{|c|}{edge2vec \cite{wang2020edge2vec}} & 0.81 & 0.80\\
        \multicolumn{1}{|c|}{UGN} & \textbf{0.82} & \textbf{0.82} \\
        \hline
        \end{tabular}}
        \caption{Social Network}
        \label{tab:SocNet_compare}
    \end{subtable}
    \begin{subtable}{.4\linewidth}
      \centering
      \scalebox{0.85}{
        \begin{tabular}{|c|ccccc|}
        \cline{2-6}
        \multicolumn{1}{l|}{}        & \multicolumn{5}{c|}{Pearson Correlation}\\ \hline
        Method & \rotatebox{90}{Res} & \rotatebox{90}{Emo} & \rotatebox{90}{Gam} & \rotatebox{90}{Lan} & \rotatebox{90}{Motor} \\
        \hline
        \cite{galan2008network} & 0.23 & 0.14 & 0.14 & 0.14 & 0.15 \\
        \cite{abdelnour2014network} & 0.23 & 0.14 & 0.14 & 0.15 & 0.15 \\
        \cite{meier2016mapping} & 0.26 & 0.16 & 0.15 & 0.16 & 0.16 \\
        \cite{abdelnour2018functional} & 0.23 & 0.14 & 0.14 & 0.15 & 0.15 \\
        NEC-DGT \cite{guo2019deep} & 0.13 & 0.34 & 0.04 & 0.17 & 0.14 \\
        GT-GAN \cite{9737289} & 0.45 & 0.34 & 0.34 & 0.35 & 0.36 \\
        UGN & \textbf{0.61} & \textbf{0.67} & \textbf{0.68} & \textbf{0.66} & \textbf{0.66} \\
        \hline
        \end{tabular}}
        \caption{HCP}
        \label{tab:HCP_compare}
    \end{subtable}%
    \begin{subtable}{.27\linewidth}
      \centering
      \scalebox{0.85}{
        \begin{tabular}{|c|c|c|c|}
        \cline{2-4}
        \multicolumn{1}{l|}{}        & \multicolumn{3}{c|}{AUPRC}\\ \hline
        Method & DDI & DTI & PPI \\
        \hline
        DeepWalk \cite{perozzi2014deepwalk} & 0.69 & 0.75 & 0.72 \\
        node2vec \cite{grover2016node2vec} & 0.79 & 0.77 & 0.77\\
        L3 \cite{kovacs2019network} & 0.85 & 0.89 & 0.90\\
        VGAE \cite{kipf2016variational} & 0.83 & 0.85 & 0.88\\
        GCN \cite{DBLP:conf/iclr/KipfW17} & 0.84 & 0.90 & 0.91\\
        SkipGNN \cite{huang2020skipgnn} & 0.85 & 0.93 & 0.92\\
        HOGCN \cite{kishan2021predicting} & \textbf{0.88} & \textbf{0.94} & 0.93 \\
        UGN & 0.85 & \textbf{0.94} & \textbf{0.96}\\
        \hline
        \end{tabular}}
        \caption{DDI, DTI and PPI}
        \label{tab:DDI_DTI_compare}
    \end{subtable}%
\caption{Performance comparison on (a) IoT, (b) Chemistry Reaction, (c) Social Network, (d) Community Detection, (e) HCP, and (f) DDI, DTI and PPI datasets}
\end{table*}

Table~\ref{tab:IoT_compare}\footnote{Best results are shown in bold-faced in this table and all other result tables.} presents performance comparison (in terms of accuracy) between our proposed framework and other novel baselines and SOTA models on the IOT dataset. We can observe that our proposed framework outperforms the baselines and produces SOTA performance on the IoT-40 and IoT-60 datasets for both node classification and edge classification tasks. For the IoT-20 dataset, our model yields SOTA results on the edge classification task and produces comparable performance on the node classification task. Our model performs better as the order of the graphs increases. However , for smaller graphs with more training data and bigger batch size, it can perform better (e.g., With 160 graph-pairs as training set, batch size 16,183 graph-pairs as test set and LR 0.005, our model yields SOTA result for the node classification task on thee IoT-20 dataset with Node-Acc. 0.93).




\subsection{Chemistry Reaction}
The chemistry reaction prediction task is divided into two phases. The first phase consists of a binary classification on atoms (nodes) predicting whether any bond (edge) attached to an atom is changed from reactant (source graph) to product (target graph). In the second phase, finding the bond types (no bond, single bond, double bond, triple bond, and aromatic bond) among the atoms whose state is predicted as changed in the first phase, is formulated as a multi-class classification problem. This two-phase approach helps to reduce total number of predictions from ${}^NC_2$ to $\left(N+{}^{N_a}C_2\right)$, where $N$ is total number of atoms and $N_a$ is the number of atoms predicted as actively participating in the reaction. When $N_a$ $<$ $N$, it leads to a reduced computational cost. To conduct the training operation, we initialize the node features with 82 dimensional vectors provided in the dataset.

Let there be $N$ nodes in the source graph and the latent dimension be $L$. The latent vector of $node_i$ be $l_i=[l_{i_0}, l_{i_1}, \dots , l_{i_{L-1}}]$. Additionally, another vector $b_i=[b_{i0}, b_{i1}, \dots , b_{i(N-1)}]$ is introduced, to incorporate the bond-information between $node_i$ and $node_j$. $b_{ij}=\frac{1}{4}$ for single bond, $\frac{2}{4}$ for double bond, $\frac{3}{4}$ for triple bond, 1 for aromatic bond, and 0 for no bond. Finally, we concatenate the node representation as well as the bond information and as a result an intermediate matrix representation ($M$) is produced of size $(L+N)\times (L+N)$ which is used as the input to the decoder.



Out of total 7,180 reactant-product graph pairs (cf. Section  \ref{dataset} Para \ref{data:chem}), we employ 1,280 data instances for training our framework and the rest of the paired instances are used for evaluation. Table~\ref{tab:Chem_compare} presents the performance comparison of our framework with the existing baselines on this dataset. We can observe that our proposed approach produces comparable performance in comparison to the SOTA models.

\subsection{Community Detection}

Community detection is a node-level multi-class classification task. Since it is a semi-supervised task, the required loss function in the training process has an unsupervised loss component ($L_u$) as mentioned in Section ~\ref{unsupervised} (cf. Equation (\ref{eqn:unsup_loss})) along with the categorical cross-entropy loss ($L_s$, cf. Equation (\ref{eqn:sup_loss})) of the nodes (one from each community) whose label (community) is revealed during the training process. We carried out training for this task by using a learning rate of 0.005 employing a combination of the unsupervised and supervised losses as mentioned in Section \ref{sec:SpecialCases}.

We carried out experiments on 3 popular Community Detection datasets (cf. Section \ref{dataset} Para \ref{data:community}) and Table \ref{tab:ComDec_compare} compares the performance of our model on these three datasets
against well-known methods on this task. Our model produces SOTA results on all three datasets on community detection.

\subsection{Social Networks}
Another important and challenging problem in graph learning is link prediction on Social Network Analysis datasets. It is essentially an edge-level binary classification task, where existence of edge between two nodes (along with its direction) is to be predicted.

We carried out (directed) link prediction on two widely used Social Network datasets -- Epinions and Slashdot.
In each dataset, number of negative edges sampled were same as the number of positive edges, in both training and test set. Due to limited RAM-size, loss of entire positive or negative edge set could not be back propagated in the training epochs. Therefore, in each epoch, 14,250 random samples were selected from each of the positive and negative edges for back propagation of loss. This sample-size may vary with available RAM-size. 
\subsubsection{Epinions}
\label{sec:epns}
For the Epinions  dataset,
out of total 355,217 positive edges, 255,217 positive edges were used as the training set and the rest (100,000 positive edges) were used for testing. Here negative edges denote neither trust nor distrust between two nodes (users). 
Using balanced negative sampling, training was conducted with a learning rate of 0.005 on total 510,434 edges and evaluation was carried out on a total of 200,000 edges. The required node features are initialized using the user ratings of the products across 27 different categories, as well as the helpfulness scores associated with those ratings. Let us consider the rating vector for $user_n$ denoted by $r^n=[r_0^n, r_1^n, \dots , r_{26}^n]$, wherein $r_i^n$ represents the total rating assigned by $user_n$ to the products falling under $category_i$. Similarly, let us denote the helpfulness vector for $user_n$ by $h^n=[h_0^n, h_1^n, \dots , h_{26}^n]$, where $h_i^n$ represents the total helpfulness of the rating $r_i^n$. The initial 54 dimensional node feature of the user can be expressed as $f^n=[\frac{r_0^n}{c_0^n}, \frac{r_1^n}{c_1^n}, \dots , \frac{r_{26}^n}{c_{26}^n}, \frac{h_0^n}{c_0^n}, \frac{h_1^n}{c_1^n}, \dots , \frac{h_{26}^n}{c_{26}^n}]$, where $c_i^n$ is the total number of ratings given by $user_n$ to the products belonging to $category_i$. It is to be noted that if no rating is assigned to $category_i$ by $user_n$, then we set $c_i^n=\epsilon$ to avoid division by zero.
Table~\ref{tab:SocNet_compare}, which presents a comparison (in terms of accuracy) against other baselines on the Social Network datasets, shows that our model produces SOTA performance on the Epinions dataset for directed link prediction task. 

  
\subsubsection{Slashdot}
\label{sec:slashdot}
Out of total 870,161 positive edges in this Slashdot dataset, 
we considered  670,161 positive edges for the training set, and the rest (200,000) for testing. Here negative edges denote neither friendship nor enmity between two nodes (users). Considering a balanced negative sampling, training was conducted with a learning rate of 0.005 on 1,340,322 edges and testing was performed on 400,000 edges. Due to the unavailability of the required node feature representation for this dataset, we utilize the supernode feature using the following strategy to generate the node feature representation.


\begin{enumerate}
\item \textbf{Supernode Creation}: As mentioned in Section~\ref{super}, 83 supernodes are constructed, the first 82 each consisting of 1000 nodes and the last one containing 168 nodes.


\item \textbf{Calculatation of Edge-Count Vectors}: Let, the incoming edge-count vector for $node_i$ be $[ic_0^i, ic_1^i, \dots , ic_{82}^i]$ where $ic_k^i$ is the total number of edges incoming from $super$-$node_k$ to $node_i$. Similarly, the outgoing edge-count vector for $node_i$ be $[og_0^i, og_1^i, \dots , og_{82}^i]$ where $og_k^i$ is the total number of edges outgoing from $node_i$ to $super$-$node_k$. Finally, the edge-count vector for $node_i$, $ec_i$, is formed by concatenating the outgoing edge-count vector with the incoming edge-count vector and normalizing the counts by the total number of nodes in the super-nodes, i.e., $ec_i=\left[\frac{ic^i_0}{1000}, \frac{ic^i_1}{1000}, \dots , \frac{ic^i_{82}}{168}, \frac{og^i_0}{1000}, \frac{og^i_1}{1000}, \dots , \frac{og_{82}^i}{168}\right]$.
\item \textbf{Feature Vectors Creation}: Final feature-vector of $node_i$, $f_i$, is obtained by augmenting the edge-count vector $ec_i$ with a 10 dimensional random-vector $r^i$ such that

$f_i=\left[\frac{ic^i_0}{1000}, \dots , \frac{ic^i_{82}}{168}, \frac{og^i_0}{1000}, \dots , \frac{og^i_{82}}{168}, r^i_0, \dots , r^i_9\right]$,$r^i_j\in[0,1] \forall j\in[0,9]$. The random-vector is introduced to ensure different feature vectors for different nodes in case the edge-count vectors of two or more nodes are the same. Thus, the initial feature-vectors of the nodes are 176 dimensional.
\end{enumerate}

Table \ref{tab:SocNet_compare} presents the results on the 2 Social Network datasets and it shows that our proposed framework achieves SOTA accuracy on both the datasets.

\subsection{Biomedical}

\subsubsection{Human Connectome Project}

For each of the five tasks in the HCP dataset (cf. Section \ref{dataset} para \ref{data:hcp}), SC and FC pairs of 239 subjects were
taken as the training set and the
rest (100 subjects) were considered as the test set. Training was performed with a batch size of 1 and a learning rate of 0.005. For each task, an MTCM representation is formed by averaging the functional connectivity (FC) matrices in the training set (cf. Section \ref{mtcm}). For each task, for each subject, a difference matrix representation is computed by subtracting the corresponding MTCM representation from its own FC representation. The model learns to predict the difference matrix from the SC information. The predicted difference matrix representation is added to the corresponding MTCM to arrive at the predicted FC representation.

Table~\ref{tab:HCP_compare} 
presents a comparative performance analysis (in terms of Pearson Correlation) of our framework with other baseline models for the SC to FC prediction task on the HCP dataset. Table \ref{tab:HCP_compare} reveals that our model outperforms all the other baseline models by a significant margin, thereby positioning itself as the SOTA technique for each of the 5 tasks 
on the HCP dataset.

\subsubsection{DDI, DTI, PPI}
\label{sec:ddi}

Table \ref{tab:DDI_DTI_PPI_expt_res} presents the experimental setup for the experiments on the BioSNAP-DDI, BioSNAP-DTI and HuRI-PPI datasets.
Table \ref{tab:DDI_DTI_compare} presents a comparison of the performance evaluation of our model, in terms of Area Under Precision Recall Curve (AUPRC), against other well-known baseline methods on the BioSNAP-DDI, BioSNAP-DTI and HuRI-PPI datasets. The results depicted in Table \ref{tab:DDI_DTI_compare} demonstrate that our model produces SOTA results for the link prediction task on BioSNAP-DTI and HuRI-PPI datasets and it performs comparably on the BioSNAP-DDI dataset.

\begin{table*}[htbp]
\centering
\scalebox{0.8}{
\begin{tabular}{||c|c|c|c|c|c|c|c|c||}
\hline
\multicolumn{1}{||l|}{Dataset} &
  \begin{tabular}[c]{@{}c@{}}\# Positive\\ Train Edges\end{tabular} &
  \begin{tabular}[c]{@{}c@{}}\# Positive\\ Test Edges\end{tabular} &
  \begin{tabular}[c]{@{}c@{}}Negative \\ Sampling ratio\end{tabular} &
  \begin{tabular}[c]{@{}c@{}}SuperNode\\ Type\end{tabular} &
  \begin{tabular}[c]{@{}c@{}}Order of \\ SuperNode\end{tabular} &
  \begin{tabular}[c]{@{}c@{}}\# SuperNodes\end{tabular} &
  \begin{tabular}[c]{@{}c@{}}SuperNode Feature\\ Dimension\end{tabular} &
  \begin{tabular}[c]{@{}c@{}}Random Vector \\ Dimension\end{tabular} \\ \hline\hline
\multirow{2}{*}{DDI} &
  \multirow{2}{*}{23,514} &
  \multirow{2}{*}{25,000} &
  \multirow{2}{*}{1:1} &
  \multirow{2}{*}{Drug} &
  101 &
  14 &
  \multirow{2}{*}{15} &
  \multirow{2}{*}{135} \\ \cline{6-7}
 &
   &
   &
   &
   &
  100 &
  1 &
   &
   \\ \hline\hline
\multirow{4}{*}{DTI} &
  \multirow{4}{*}{10,638} &
  \multirow{4}{*}{4,500} &
  \multirow{4}{*}{1:1} &
  \multirow{2}{*}{Drug} &
  66 &
  75 &
  \multirow{4}{*}{111} &
  \multirow{4}{*}{59} \\ \cline{6-7}
 &
   &
   &
   &
   &
  67 &
  1 &
   &
   \\ \cline{5-7}
 &
   &
   &
   &
  \multirow{2}{*}{Target} &
  66 &
  21 &
   &
   \\ \cline{6-7}
 &
   &
   &
   &
   &
  67 &
  14 &
   &
   \\ \hline\hline
\multirow{2}{*}{PPI} &
  \multirow{2}{*}{36,048} &
  \multirow{2}{*}{15,620} &
  \multirow{2}{*}{1:1} &
  \multirow{2}{*}{Target} &
  81 &
  5 &
  \multirow{2}{*}{103} &
  \multirow{2}{*}{57} \\ \cline{6-7}
 &
   &
   &
   &
   &
  80 &
  98 &
   &
   \\ \hline
\end{tabular}
}
\caption{Experimental setup for DDI, DTI and PPI datasets}
\label{tab:DDI_DTI_PPI_expt_res}
\end{table*}

\begin{table*}[htbp]
    \begin{minipage}{.3\linewidth}
      \centering
      \begin{tabular}{||c|c||}
        \hline
        Method & HITS@1\\
        \hline\hline
        \makecell{DihEdral \cite{xu2019relation}} & 0.38\\
        \makecell{InteractE \cite{vashishth2020interacte}} & 0.46\\
        \makecell{RefE \cite{chami2020low}} & 0.50\\
        \makecell{MEI \cite{nghiep2020multi}} & 0.51\\
        \makecell{ComplEx-DURA \cite{zhang2020duality}} & 0.51\\
        \makecell{MEIM \cite{DBLP:conf/ijcai/TranT22}} & 0.51\\
        UGN & \textbf{0.61}\\
        \hline
        \end{tabular}
        \caption{Performance Comparison on the YAGO dataset}
        \label{tab:YAGO_compare}
    \end{minipage}%
    \begin{minipage}{.3\linewidth}
    \centering
    \begin{tabular}{||c|c|c|c||}
        \hline
        Train & Test & AUPRC & Acc.\\
        \hline\hline
        \rowcolor{Gray}
        DDI & DDI & 0.85 & 0.78\\
        PPI & DDI & \textit{\textbf{0.86}} & \textit{0.78}\\
        DTI & DDI & 0.82 & \textit{\textbf{0.80}}\\
        \hline\hline
        \rowcolor{Gray}
        PPI & PPI & 0.96 & 0.90\\
        DDI & PPI & \textit{\textbf{0.97}} & \textit{0.91}\\
        DTI & PPI & \textit{\textbf{0.97}} & \textit{\textbf{0.92}}\\
        \hline\hline
        \rowcolor{Gray}
        DTI & DTI & \textbf{0.94} & \textbf{0.84}\\
        DDI & DTI & 0.92 & \textit{\textbf{0.84}}\\
        PPI & DTI & 0.90 & 0.82\\
        \hline
        \end{tabular}
        \caption{Zero-Shot learning on \\the biomedical datasets}      \label{tab:biomedical_zeroshot}
    \end{minipage}%
    \begin{minipage}{.3\linewidth}
    \centering
    \begin{tabular}{||c|c|c||}
        \hline
        \makecell{Max. train samples\\per relation type} & \makecell{Test\\accuracy} & \makecell{Number of\\train edges}\\
        \hline\hline
        10 & 0.10 & 370\\
        50 & 0.14 & 1,819\\
        100 & 0.16 & 3,587\\
        500 & 0.28 & 16,504\\
        1,000 & 0.36 & 30,656\\
        2,000 & 0.42 & 55,958\\
        4,000 & 0.48 & 99,351\\
        \hdashline
        8,000 & 0.53 & 156,184\\
        16,000 & 0.57 & 224,142\\
        \hline
        \end{tabular}
        \caption{Few-Shot learning on the YAGO dataset}
        \label{tab:YAGO_fewshot}
    \end{minipage}%
\end{table*}


\subsection{Knowledge Graph Completion}
Knowledge Graph (KG) Completion is a fundamental problem in the field of artificial intelligence and natural language processing. It involves predicting missing relationships or facts in a KG, which is a structured representation of information as interconnected nodes and edges.  

For the experiment on the YAGO dataset, 1,079,040 relations (directed edges between entity-pairs) were used as the training set with sample-size 370 (10 samples from each type of relation) per epoch with an LR of 0.005, and 10,000 edges were used as the test set. For initializing the node features, following steps are used for each entity $e$.
\begin{enumerate}
\item A 37 dimensional vector $L_e$ named ``Left-Entity-One-Hot" is created in such a way that,

\vspace*{-\baselineskip}
\[
    L_e[i]= 
\begin{cases}
    1,              & \text{if } \exists \text{ relation } r_i \text{ and entity } k \text{ so that,}\\
    & \text{relationship triplet }(e,r_i,k) \in \text{training set}\\
    0,              & \text{otherwise}
\end{cases}
\]
\item A 37 dimensional vector $R_e$ named ``Right-Entity-One-Hot" is created in such a way that,
\[
    R_e[i]= 
\begin{cases}
    1,              & \text{if } \exists \text{ relation } r_i \text{ and entity } k \text{ so that,}\\
    & \text{relationship triplet }(k,r_i,e) \in \text{training set}\\
    0,              & \text{otherwise}
\end{cases}
\]
\item Concatenate $L_e$, $R_e$, and a 20 dimensional random vector whose values lie between 0 and -1 making the initial node features of size 94.
\end{enumerate}
Table \ref{tab:YAGO_compare} compares the performance of our model on the YAGO dataset in terms of edge accuracy or HITS@1 to other well-known methods. Table \ref{tab:YAGO_compare} shows that our model outperforms the existing models by large margins and produces SOTA results in the knowledge graph completion task on the YAGO dataset.



\section{Performance in Zero-Shot and Few-Shot Setup}
Conducting domain-specific training from scratch, using sufficient number of annotated samples, is time-consuming and resource-intensive. Zero-shot or few-shot setups play an important role in such cases, particularly in low-resource scenario.
We investigated the performance of our model in zero-shot or few-shot setups. Zero-shot experiments were carried out on three biomedical datasets -- DDI, DTI and PPI, to exploit the similarity of the three domains. In the zero-shot experiments, learning and evaluation were carried out on different datasets. For the zero-shot set up, initial node feature dimensions of the test-domain nodes are changed by changing the number of supernodes or changing the dimension of the random vector to turn the test-domain node features with dimensions equal to that of the train-domain node features. If the dimension is to be increased, then the number of supernodes is increased, and if the dimension is to be decreased, then the random vector dimension is reduced. The results of the zero-shot experiments are reported in Table \ref{tab:biomedical_zeroshot} in terms of AUPRC and edge-accuracy. In order to compare the results of the zero-shot experiments with the in-domain experiments, we also include the results of the in-domain experiments (cf. Table \ref{tab:DDI_DTI_compare}) in Table \ref{tab:biomedical_zeroshot} (shaded rows). Table \ref{tab:biomedical_zeroshot} shows that in most of the cases, Zero-Shot learning in cross-domain produces better results than the original (in-domain) experiments. We observe that, if in one domain, when the graph is bipartite (e.g., DTI dataset) and in the other domain, the graph is single or unpaired (e.g., DDI or PPI dataset) the cross-domain experiments typically does not produce better results in terms of AURPC score. However, it is interesting to note that for the model trained on DTI and tested on DDI or PPI, the accuracy improves even though the AUPRC decreases. The italicized scores in Table \ref{tab:biomedical_zeroshot} are same or better than the in-domain experiment scores.

Few-Shot learning was performed on the YAGO dataset. Maximum number of samples per relation type was increased step by step to observe how much data is required for the model to perform better than the
SOTA model and the results are presented in Table \ref{tab:YAGO_fewshot}. Table \ref{tab:YAGO_fewshot} shows that, with our method only 156,184 edges out of 1,079,040 edges in the training set with maximum training samples of 8,000 per relation type is adequate to beat the existing SOTA results. 

\section{Ablation Study}

\begin{figure}[t]
\centering
\vspace{-.2cm}
\includegraphics[width=0.5\textwidth]{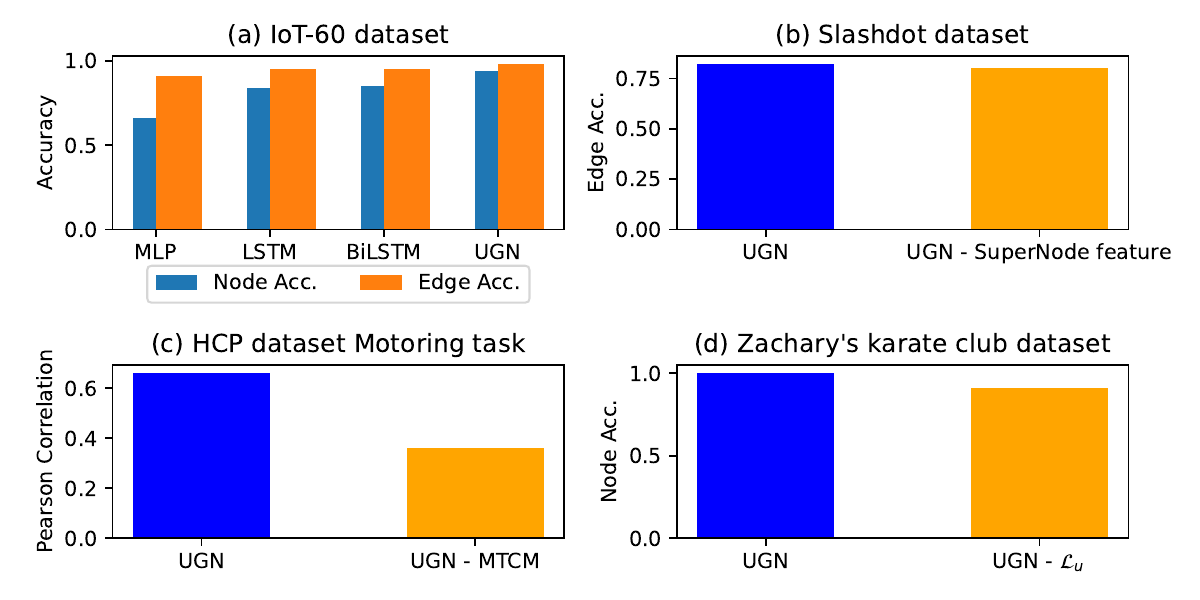}
\caption{
\small Ablation study by component replacement and removal on (a) IoT-60 dataset, (b) Slashdot dataset, (c) HCP dataset Motoring task, and (d) Zachary's karate club dataset
}
\label{fig:ablation}
\end{figure}
To examine the impact of the intermediate matrix representation and the convolution layer-based decoder on the overall performance of the model, we conduct an ablation study using the IoT-60 dataset.
We follow the same pathway of existing encoder-decoder based graph neural network literature, where the output representation from the encoder module is passed to the decoder module without forming an additional intermediate matrix representation for the decoder module. More specifically, latent vector representation of the nodes from the encoder block are passed as input to the decoder for the node classification task. For the edge classification, initially an edge vector \(e_{i,j}\) is formed by element-wise multiplication of the latent vectors (\(l_i, l_j\)) of the two nodes for each edge, i.e., 
$e_{i,j} = [l_{i_0} \cdot l_{j_0}, l_{i_1} \cdot l_{j_1}, \dots, l_{i_{L-1}} \cdot l_{j_{L-1}}]$,  
and is passed to the decoder as input. Additionally, to investigate the contribution of different blocks in comparison to the convolutional layers, we employ different existing neural network blocks such as multi-layer-perceptron (MLP), long-short-term-memory (LSTM) and bidirectional-LSTM (BiLSTM). The results of these experiments are reported in Figure~\ref{fig:ablation}a. 
From Figure~\ref{fig:ablation}a, we can observe that  UGN outperforms other models by significant margins in terms of accuracy on the IoT-60 dataset.

To assess the impact of the supernode based features, we conducted an experiment on the Slashdot dataset with randomly initialized node features having values between 0 and 1. All other necessary hyper-parameters were kept the same to measure the contribution of the supernode based features. From Figure~\ref{fig:ablation}b, 
we can observe an edge accuracy of 0.80 using this setup, i.e., without utilizing the supernode features, which is the same as obtained with edge2vec~\cite{wang2020edge2vec} (cf. Table \ref{tab:SocNet_compare}).
Thus our baseline model without the supernode features performs at par with edge2vec~\cite{wang2020edge2vec} on the directed link prediction task on the Slashdot dataset and the supernode based features bring in the additional improvements.

To examine the significance of MTCM for complete graphs, we conducted another experiment in which the MTCM component was removed from the training process. 
We modified the training procedure of the model on the HCP dataset, where the model learned to predict the FC matrix directly from the SC matrix of a subject. 
In this experimental setup, we achieved an average pearson correlation of 0.36 for the motoring task on the HCP dataset, as mentioned in Figure~\ref{fig:ablation}c, keeping all other parameters and hyper-parameters the same. 
Thus, our baseline architecture without the MTCM component performs at par with the previous SOTA model, GT-GAN \cite{9737289}, on the HCP dataset. Introduction of the MTCM module into our framework helps in significant performance gain and producing SOTA performance.  



Finally, we conduct another ablation experiment to study the effect of the unsupervised component ($\mathcal{L}_u$) of the loss function (cf. equation \ref{eqn:unsup_loss}) for semi-supervised learning tasks. To carry out the experiment, we removed $\mathcal{L}_u$ from the final loss function ($\mathcal{L}$). Keeping everything else unchanged, we conducted the ablation experiment on the Zachary's karate club dataset. Our empirical findings from Figure~\ref{fig:ablation}d demonstrate that the unsupervised component ($\mathcal{L}_u$) of the final loss function ($\mathcal{L}$) helps significantly in achieving state-of-the-art (SOTA) performance on Zachary's karate club dataset for the semi-supervised learning task.



\section{Conclusion}
In this paper, we presented a thorough investigation of the feasibility of applying encoder-decoder based frameworks for graph analysis tasks by providing valuable insights into its applications, challenges and potential solutions. Through an in-depth analysis, we can conclude concretely that our novel UGN framework performs relatively well by producing SOTA results on few datasets and produce comparable results on the rest of the datasets. In this work, we have developed a novel generalized framework and evaluated it on six different tasks by employing twelve datasets. We have introduced some task specific add-ons like supernode feature for graphs, where node feature is not available, unsupervised loss component for semi-supervised settings, and MTCM for complete graphs. Our extensive empirical analysis suggests that our proposed UGN produces SOTA results on ten datasets and comparable results on the remaining two datasets. Additionally, our work has contributed to the field of graph learning and analysis by providing a comprehensive overview of the existing state-of-the-art techniques, identifying challenges, and proposing novel solutions. The findings of this research serves as a valuable resource for researchers, practitioners, and stakeholders interested in leveraging the power of graph data for various applications, e.g., Supply Chain Optimization, Fraud Detection, Recommendation Systems, Cyber-security, etc. As the field continues to evolve, future research has the opportunity to extend the foundations established in this work, thereby advancing the comprehension, scalability, and applicability of graph learning and analysis techniques.



\bibliography{ref}


\end{document}